\documentclass[10pt,twocolumn,letterpaper]{article}

\usepackage{cvpr}
\usepackage{times,flushend}
\usepackage{graphicx}
\usepackage{amsmath}
\usepackage{amssymb}
\usepackage[export]{adjustbox}
\usepackage{booktabs}

\usepackage{fancyheadings}
\usepackage[pagebackref=true,breaklinks=true,letterpaper=true,colorlinks,bookmarks=false]{hyperref}

\cvprfinalcopy 

\begin{document}

\title{Classification and Retrieval of Digital Pathology Scans: A New Dataset}

\author{Morteza Babaie$^{1,2}$, Shivam Kalra$^{1}$, Aditya Sriram$^{1}$, Christopher Mitcheltree$^{1,3}$, \\Shujin Zhu$^{1,4}$, Amin Khatami$^{5}$, 
Shahryar Rahnamayan$^{1,6}$, H.R. Tizhoosh$^{1}$\vspace{0.1in}\\
$^1$ KIMIA Lab, University of Waterloo, Canada\\
$^2$ Mathematics and Computer Science, Amirkabir University, Tehran, Iran\\
$^3$ Electrical and Computer Engineering, University of Waterloo, Canada \\
$^4$  School of Electronic \& Optical Eng., Nanjing University of Science \& Technology, Jiangsu, China \\
$^5$ Institute for Intelligent Systems Research and Innovation, Deakin University, Australia \\
$^6$ Electrical, Computer and Software Engineering, \\ University of Ontario Institute of Technology, Oshawa, Canada
}

\maketitle

\begin{abstract}
In this paper, we introduce a new dataset, \textbf{Kimia Path24}, for image classification and retrieval in digital pathology. We use the whole scan images of 24 different tissue textures to generate 1,325 test patches of size 1000$\times$1000 (0.5mm$\times$0.5mm). Training data can be generated according to preferences of algorithm designer and can range from approximately 27,000 to over 50,000 patches if the preset parameters are adopted. We propose a compound patch-and-scan accuracy measurement that makes achieving high accuracies quite challenging. In addition, we set the benchmarking line by applying LBP, dictionary approach and convolutional neural nets (CNNs) and report their results. The highest accuracy was 41.80\% for CNN.
\end{abstract}
\section{Introduction}

The integration of algorithms for classification and retrieval in medical images through effective machine learning schemes is at the forefront of modern medicine \cite{carpenter2012call}. These tasks are crucial, among others, to detect and analyze abnormalities and malignancies to contribute to more informed diagnosis and decision makings. Digital pathology is one of the domains where such tasks can support more reliable decisions \cite{Madabhushi2017}. For several decades, the archiving of microscopic information of specimens has been organized through employing and storing glass slides \cite{Janabi2011}. Beyond the fragile nature of glass slides, hospitals and clinics need large and specially prepared storage rooms to store specimens, which naturally requires a lot of logistical infrastructures. 

Digital pathology, or whole slide imaging (WSI), can not only provide high image quality that is not subject to decay (i.e., stains decay over time) but also offers a range of other benefits \cite{Janabi2011,Gabril2010}: They can be investigated by multiple experts at the same time,  they can be more easily retrieved for research and quality control, and of course, WSI can be integrated into existing information systems of hospitals. In 1999, Wetzel and Gilbertson developed the first automated WSI system \cite{ho2006use}, utilizing high resolution to enable pathologists to buffer through immaculate details presented through digitized pathology slides. Ever since, pathology bounded by WSI systems is emerging into an era of digital specialty, providing solutions for centralizing diagnostic solutions by improving the quality of diagnosis, patient safety, and economic concerns \cite{ghaznavi2013digital}. Like any other new technology, digital pathology has its pitfalls. The \emph{gigapixel} nature of WSI scans makes it difficult to store, transfer, and process samples in real-time. One also need tremendous digital storage to archive them. In this paper, we propose a new and uniquely designed data set, \emph{Kimia Path24}, for the classification and retrieval of digitized pathology images. In particular, the data set is comprised of 24 WSI scans of different tissue textures from which 1,325 test patches sized 1000$\times$1000 are manually selected with special attention to textural differences. The proposed data set is structured to mimic retrieval tasks in clinical practice; hence, the users have the flexibility to create training patches, ranging from 27,000 to over 50,000 patches -- these numbers depend on the selection of homogeneity and overlap for every given slide. For retrieval, a weighted accuracy measure is provided to enable a unified benchmark for future works.
 
\section{Related Works}
\label{sec:lit_rev}
This section covers a brief literature review on image analysis in digital pathology, specifically on WSI, followed by various content-based medical image retrieval techniques, and finally an overview of feature extraction techniques that emphasize local binary patterns (LBP). 

\subsection{Image Analysis in Digital Pathology}
In digital pathology, the large dimensionality of the image poses a challenge for computation and storage; hence, contextually understanding regions of interest of an image helps in quicker diagnosis and detection for implementing soft-computing techniques \cite{caicedo2011content}. Over the years, traditional image-processing tasks such as filtering, registration, and segmentation, classification and retrieval have gained more significance. Particularly for histopathology, the cell structures such as cell nuclei, glands, and lymphocytes are observed to hold prominent characteristics that serve as a hallmark for detecting cancerous cells \cite{gurcan2009review}. Researchers also anticipate that one can correlate histological patterns with protein and gene expression, perform exploratory histopathology image analysis, and perform computer aided diagnostics (CADx) to provide pathologists with the required support for decision making \cite{gurcan2009review}. The idea behind CADx to quantify spatial histopathology structures has been under investigation since the 1990s, as presented by Wiend et al. \cite{weind1998invasive}, Bartels et al. \cite{bartels1992bayesian}, and Hamilton et al.  \cite{hamilton1994expert}. However, due to limited computational resources and its associated expense, implementing such ideas have been overlooked or delayed. In recent years, however, WSI technology has been gradually setting laboratory standards as a process of digitizing pathology slides to advocate for more efficient diagnostic, educational and research purposes \cite{pantanowitz2011review}. This approach, unlike photo-microscopy which is to capture a portion of an image \cite{williams2010telepathology}, offers a high-resolution overview of the entire specimen in the slide which enables the pathologist to take control over navigating through the slide and saving invaluable time \cite{ho2006use, weinstein2009overview, della2009eslide, ghaznavi2013digital}. More recently, Bankhead et al.  \cite{bankhead2017qupath} provided an open-source bio-imaging software, called \emph{QuPath} that supports WSI by providing tumor identification and biomarker evaluation tools which developers can use to implement new algorithms to further improve the outcome of analyzing complex tissue images. 

\subsection{Image Retrieval}
Retrieving similar (visual) semantics of an image requires extracting salient features that are descriptive of the image content. At its entirety, there are two main points of view for processing the WSI scans \cite{barker2016automated}. The first one is called \emph{sub-setting methods} which considers a small section of the huge pathology image as an important part such that the processing of the small subset substantially reduces processing time. The majority of research in the literature prefers this method because of its advantage of speed and accuracy. However, it needs expert knowledge and intervention to extract the proper subset. On the other hand, \emph{tiling methods} break the images into smaller and controllable patches and try to process them against each other \cite{gutman2013cancer}. This naturally requires more care in design and is more expensive in execution. However, it certainly is an obvious approach toward full automation. 
\\ Traditionally, a large medical image database is packaged with textual annotations classified by specialists; however, this approach does not perform well against the ever demanding growth of digital pathology. In 2003, Zheng et al.  \cite{zheng2003design} developed an on-line content-based image retrieval (CBIR) system wherein the client provides a query image and corresponding search parameters to the server side. The server then performs similarity searches based on feature types such as color histogram, image texture, Fourier coefficients, and wavelet coefficients, whilst using the vector dot-product as a distance metric for retrieval. The server then returns images that are similar to the query image along with the similarity scores and a feature descriptor. Mehta et al. \cite{mehta2009content}, on the other hand, proposed an offline CBIR system which utilizes sub-images rather than the entire histopathology slide. Using scale-invariant feature extraction (SIFT) \cite{lowe1999object} to search for similar structures by indexing each sub-image, the experimental results suggested, when compared to manual search, an 80\% accuracy for the top-5 results retrieved from a database that holds 50 IHC stained pathology images, consisting of $8$ resolution levels. In 2012, Akakin and Gurcan \cite{akakin2012content} developed a multi-tiered CBIR system based on WSI, which is capable of classifying and retrieving scans using both multi-image query and images at a slide-level. The authors test the proposed system on $1,666$ WSI scans extracted from $57$ follicular lymphoma (FL) tissue slides containing $3$ subtypes and $44$ neuroblastoma (NB) tissue slides comprised of $4$ subtypes. Experimental results suggested a $93$\% and $86$\% average classification accuracy for FL and NB diseases, respectively. More recently, Zhang et al.  \cite{zhang2015towards} developed a scalable CBIR method to cope with WSI by using a supervised kernel hashing technique which compresses a 10,000-dimensional feature vector into only 10 binary bits, which is observed to preserve a concise representation of the image. These condensed binary codes are then used to index all existing images for quick retrieval for of new query images. The proposed framework is validated on a breast histopathology data set comprised of 3,121 WSI scans from 116 patients; experimental results state an accuracy of 88.1\% for processing at the speed of 10ms for all 800 testing images. 

\subsection{LBP Descriptor}
To generate preliminary results for the introduced data set, we captured the textural structure of patches by extracting local binary patterns (LBP) \cite{ojala1994performance} as they are among established approaches proven to quantify important textures in medical imaging \cite{nanni2010local, reyad2014comparison, icpram17}. We also experiment with the dictionary approach \cite{joachims1998text, mccallum1998comparison} and convolutional neural networks (CNN) \cite{lecun1998gradient}. 

LBP is an extremely powerful and concise texture feature extractor, with an ability to compete with state-of-the-art complex learning algorithms. In 2009, Masood and Rajpoot  \cite{masood2009texture} implemented a circular LBP (CLBP) feature extraction algorithm to classify colon tissue patterns using a Gaussian-kernel SVM on biopsy samples taken from 32 different patients. Each image has a spatial resolution of $491\times652\times128$ pixels, for which the retrieval accuracy is computed to be 90\% to distinguish between benign and malignant patterns. In the same year, Sertel et al.  \cite{sertel2009computer} presented a CADx system designed to classify Neuroblastoma (NB) malignancy, a type of cancer in the nervous system, using WSI. The authors proposed a multi-resolution LBP approach which initially analyzes image at the lowest resolution and then switches to higher resolutions when necessary. The proposed approach employs offline feature selection, which enables the extraction of more discriminative features for every resolution level during the training phase. For retrieval, a modified $k$-nearest neighbor is employed which when tested on 43 WSI scans, provides an overall classification accuracy of 88.4\%. More recently, Tashk et al. \cite{tashk2013automatic} proposed a statistical approach based on color information such as maximum likelihood estimation. Then, the CLBP is employed to extract texture features from rotational and color changes, from which the SVM algorithm classifies the extracted feature vectors as mitosis and non-mitosis cases. The proposed scheme obtains 70.94\% (F-measure) for Aperio XT images and 70.11\% for Hamamatsu images, both of which are microscopic scanners. The reported method is observed to outperform other participants at ICPR 2012 Mitosis detection in breast cancer histopathological images.

\section{The ``\emph{Kimia Path24}'' Dataset}
\label{sec:dataset}

We had 350 whole scan images (WSIs) from diverse body parts at our disposal. The images were captured by \emph{TissueScope LE 1.0}\footnote{http://www.hurondigitalpathology.com/tissuescope-le-3/}. The scans were performed in the bright field using a 0.75 NA lens. For each image, one can determine the resolution by checking the description tag in the header of the file. For instance, if the resolution is 0.5$\mu$m, then the magnification is 20x, and if the resolution is 0.25$\mu$m, then the magnification is 40x.

\begin{figure*}[htbp]
\begin{center}
\includegraphics[width=6.25cm, height=6.25cm,frame]{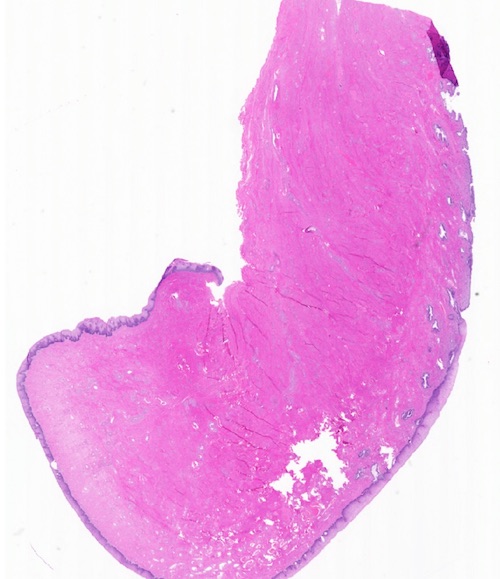}
\includegraphics[width=6.25cm, height=6.25cm,frame]{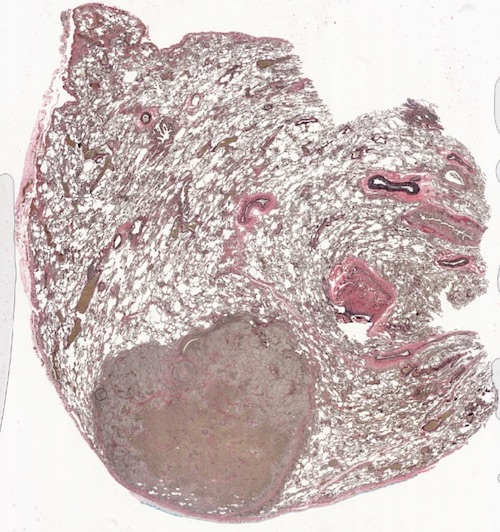}
\includegraphics[width=6.25cm, height=6.25cm,frame]{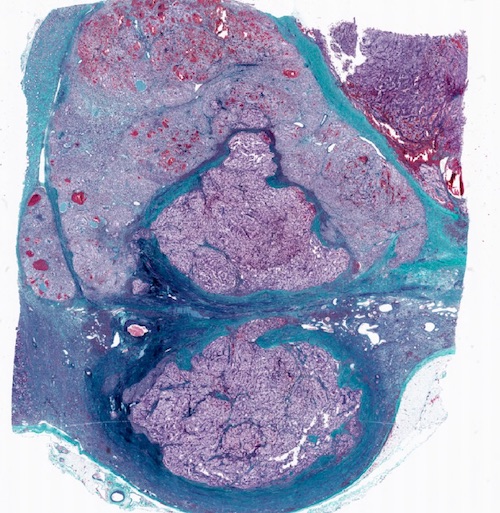}
\includegraphics[width=6.25cm, height=6.25cm,frame]{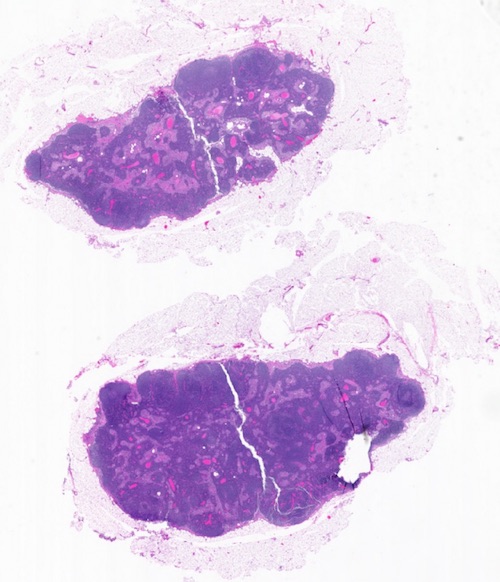}
\includegraphics[width=6.25cm, height=6.25cm,frame]{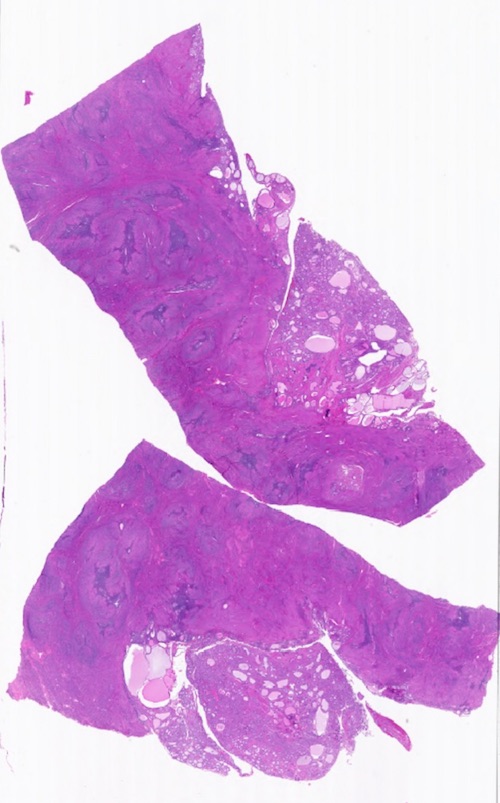}
\includegraphics[width=6.25cm, height=6.25cm,frame]{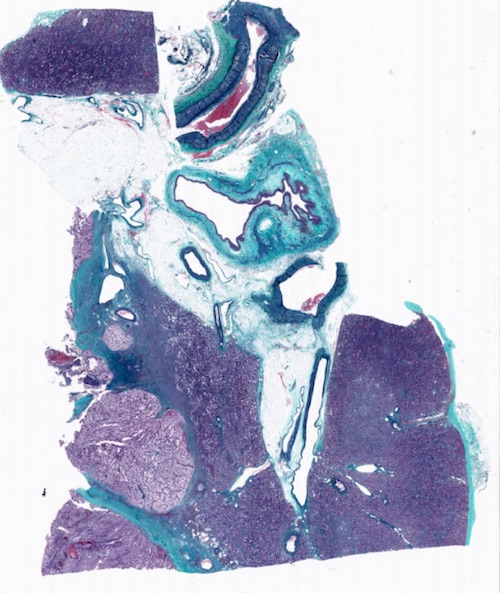}
\caption{The ``Kimia Path24'' Dataset: Thumbnails of six sample whole scan images. Aspect ratio has been neglected for better illustration.}
\label{fig:WSIthumbnails}
\end{center}
\end{figure*}

\begin{figure*}[htbp]
\begin{center}
\includegraphics[width=0.28\textwidth]{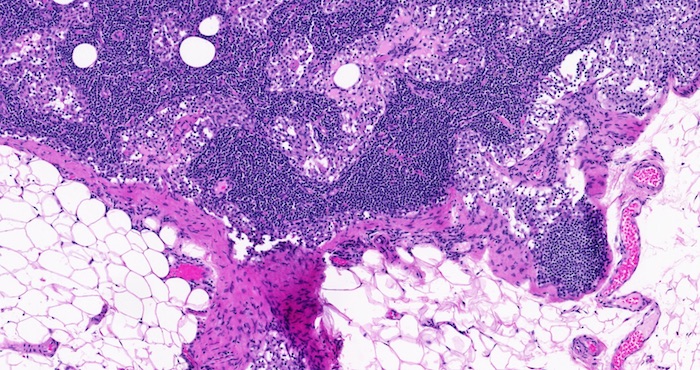}
\includegraphics[width=0.28\textwidth]{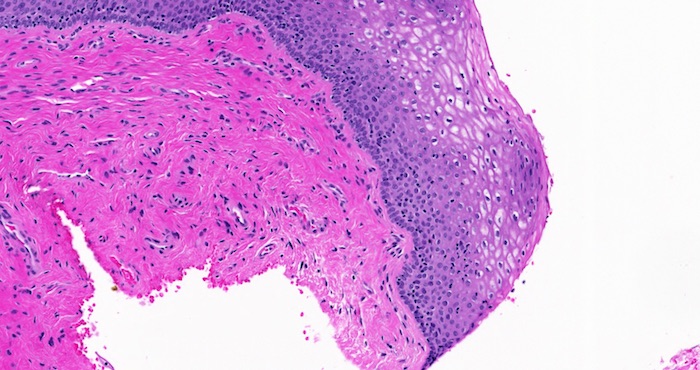}
\includegraphics[width=0.28\textwidth]{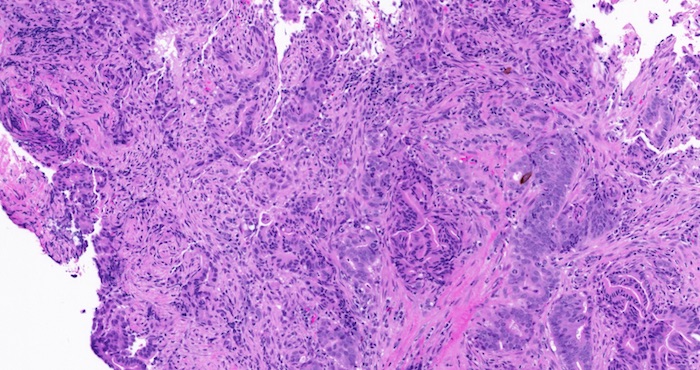}
\includegraphics[width=0.28\textwidth]{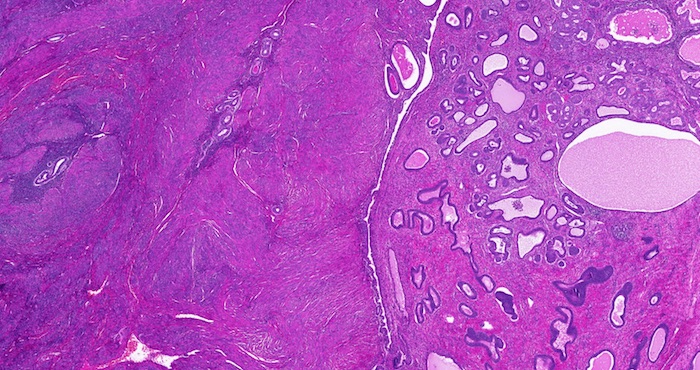}
\includegraphics[width=0.28\textwidth]{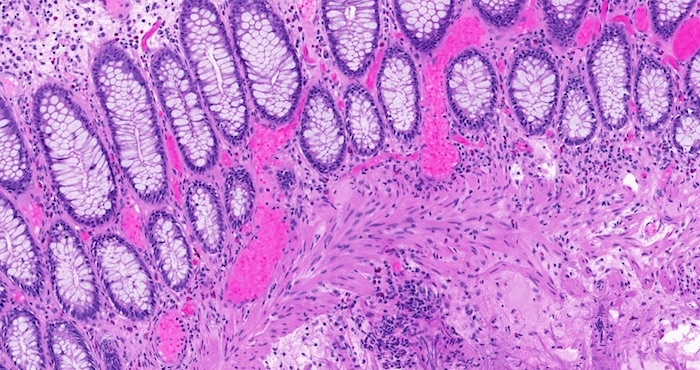}
\includegraphics[width=0.28\textwidth]{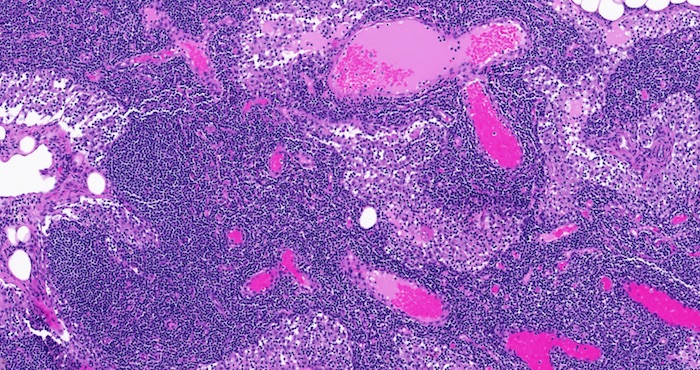}
\includegraphics[width=0.28\textwidth]{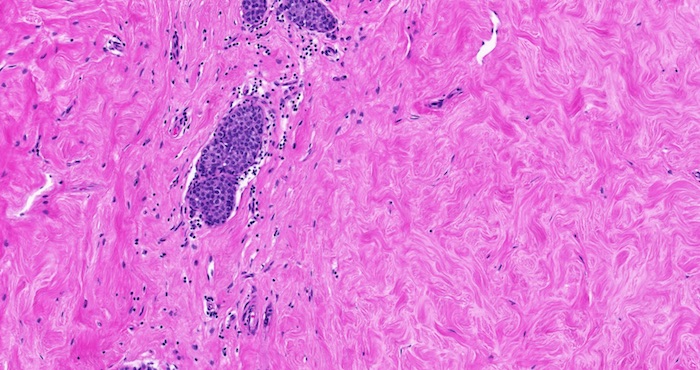}
\includegraphics[width=0.28\textwidth]{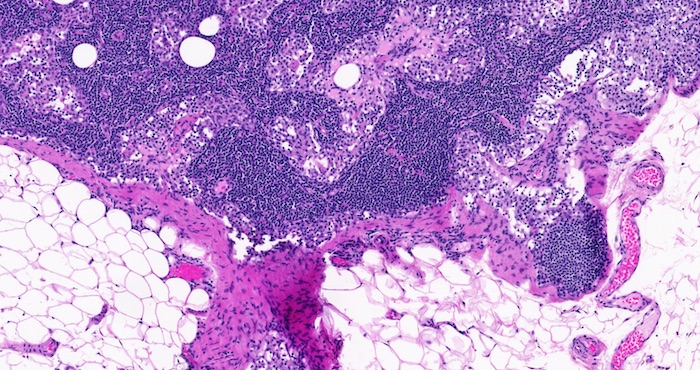}
\includegraphics[width=0.28\textwidth]{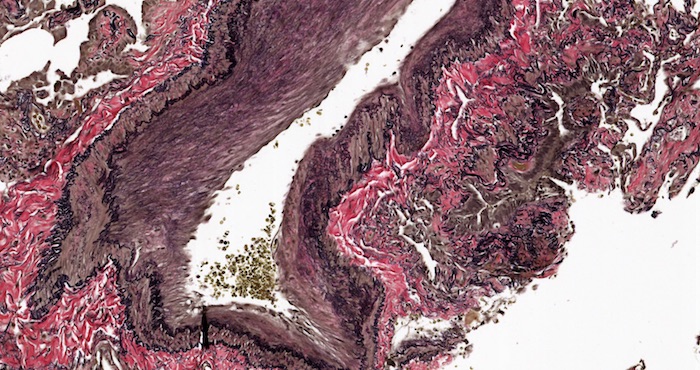}
\includegraphics[width=0.28\textwidth]{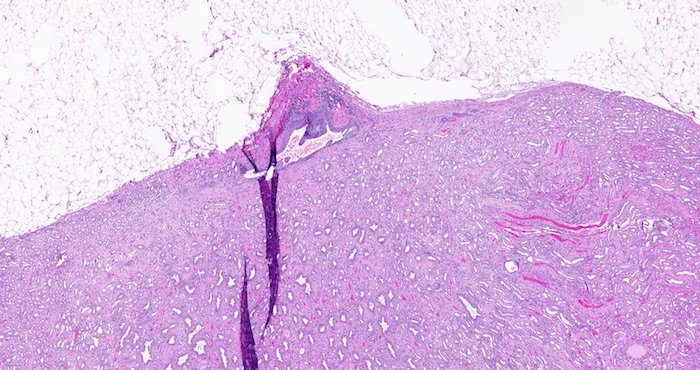}
\includegraphics[width=0.28\textwidth]{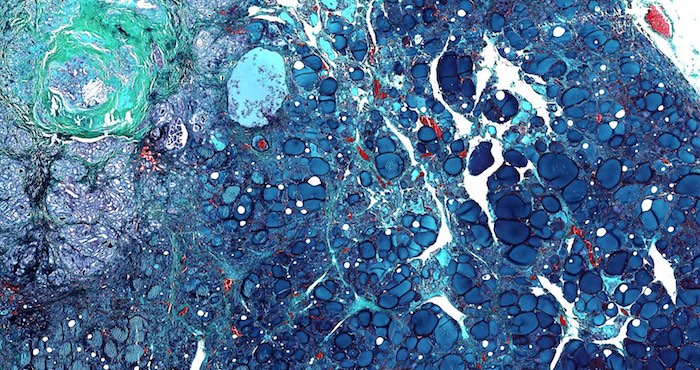}
\includegraphics[width=0.28\textwidth]{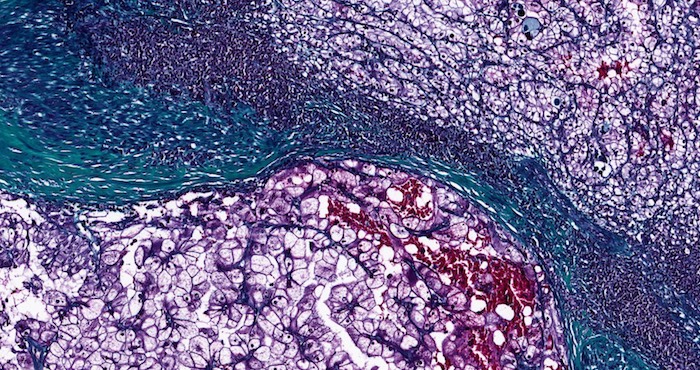}
\includegraphics[width=0.28\textwidth]{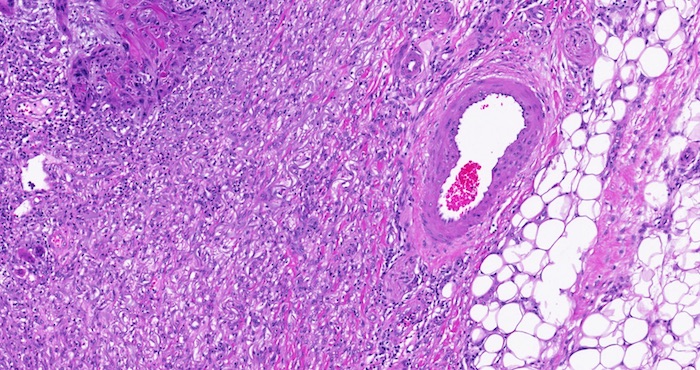}
\includegraphics[width=0.28\textwidth]{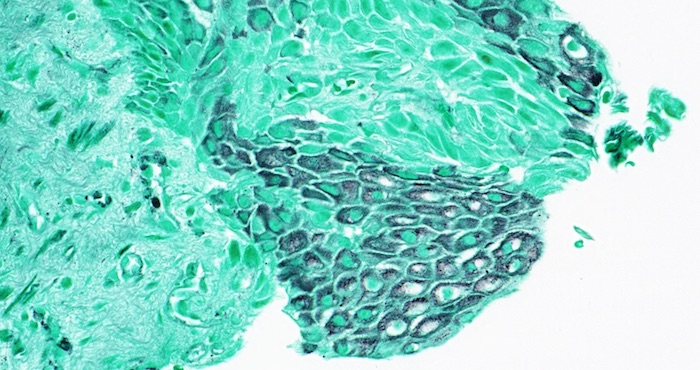}
\includegraphics[width=0.28\textwidth]{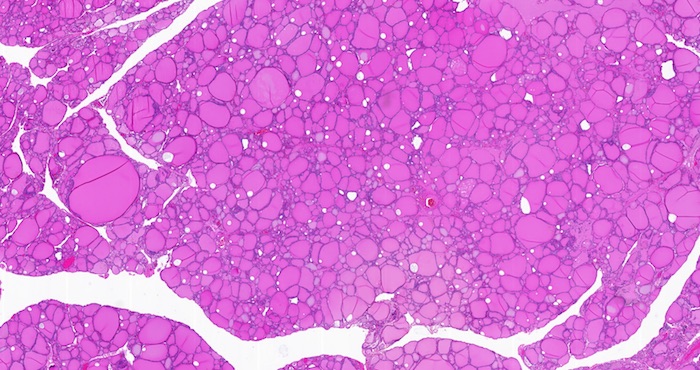}
\includegraphics[width=0.28\textwidth]{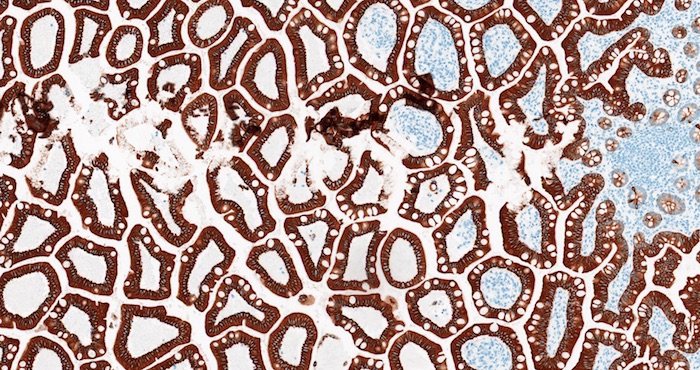}
\includegraphics[width=0.28\textwidth]{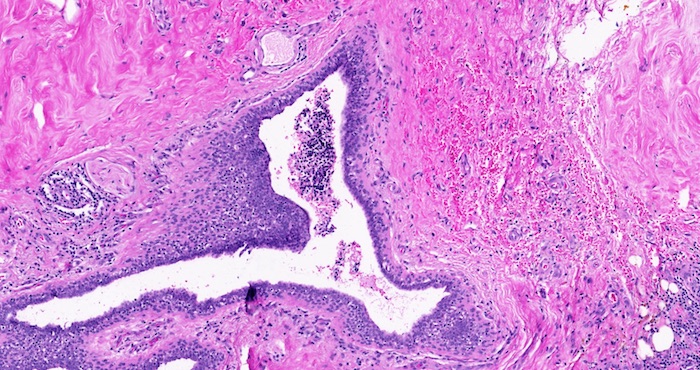}
\includegraphics[width=0.28\textwidth]{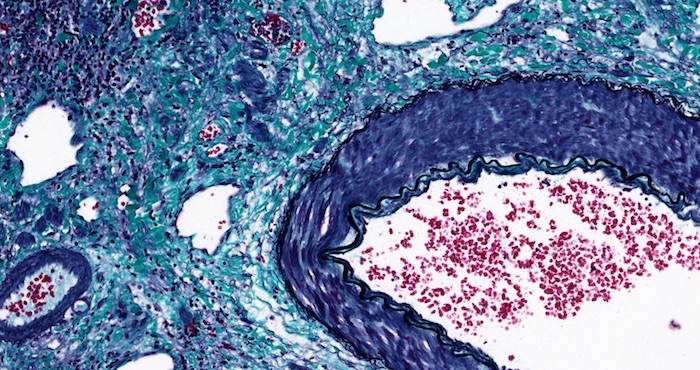}
\includegraphics[width=0.28\textwidth]{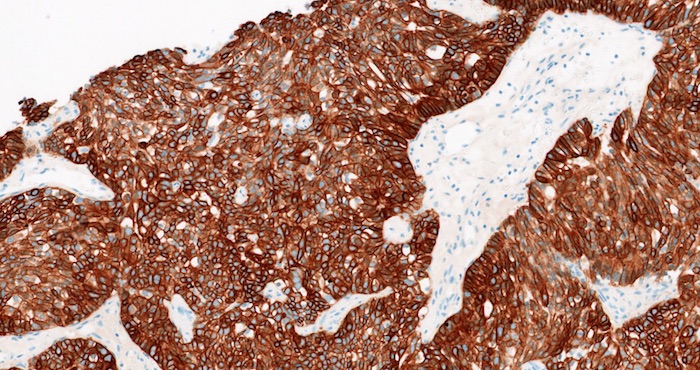}
\includegraphics[width=0.28\textwidth]{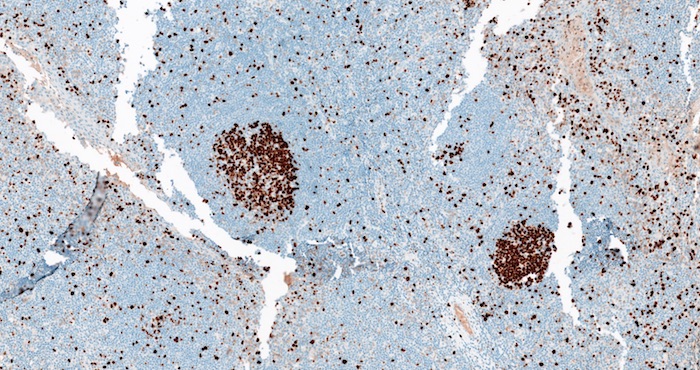}
\includegraphics[width=0.28\textwidth]{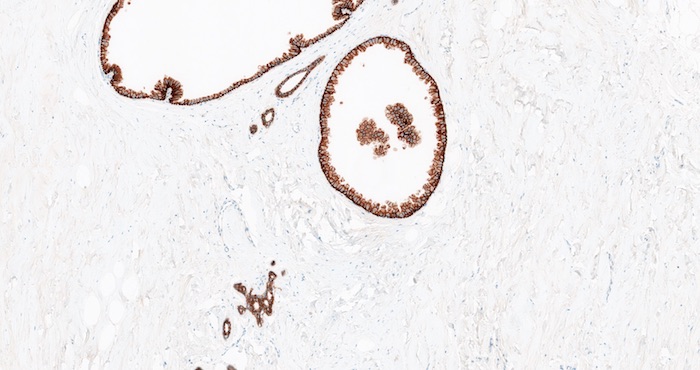}
\includegraphics[width=0.28\textwidth]{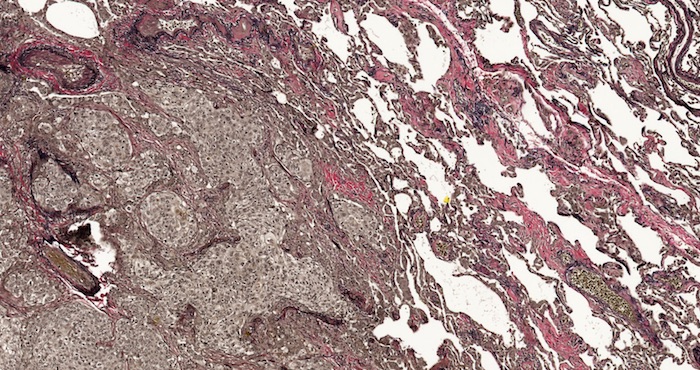}
\includegraphics[width=0.28\textwidth]{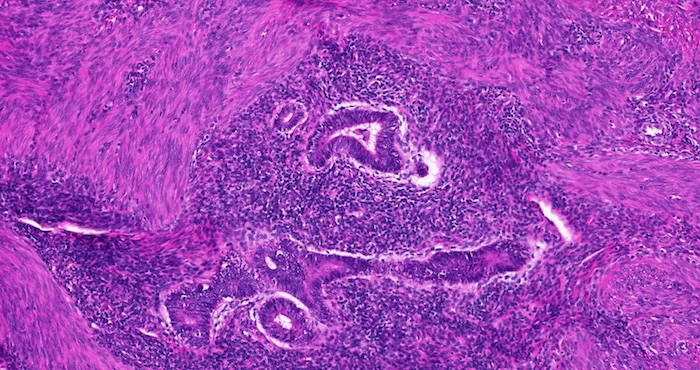}
\includegraphics[width=0.28\textwidth]{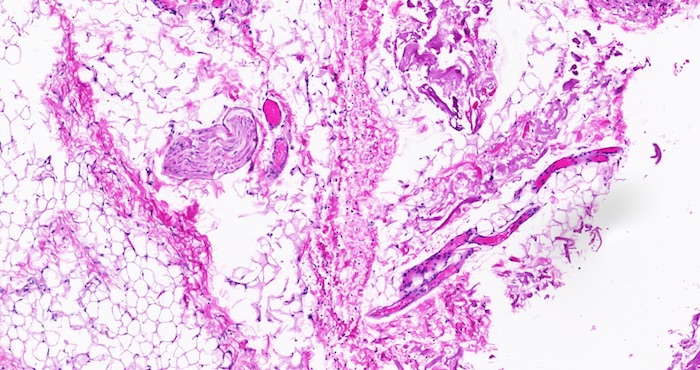}
\caption{All 24 scans used to generate the \emph{Kimia Path24} dataset. The images represent approximately 20x magnification of a portion of the whole scan images as depicted in Fig. \ref{fig:WSIthumbnails}. }
\label{fig:WSImagnified}
\end{center}
\end{figure*}

 We manually selected 24 WSIs purely based on visual distinction for non-clinical experts which means, in our selection, we made conscious effort to select a subset of the WSIs such that they clearly represent different \emph{texture} patterns. Fig. \ref{fig:WSIthumbnails} shows the thumbnails of six samples. Fig. \ref{fig:WSImagnified} displays a magnified portion of each WSI.

Our intention is to provide a fixed testing dataset to facilitate benchmarking but respect the design freedom of individual algorithm designer to generate his own training dataset. To achieve this, we performed the following steps:

\begin{enumerate}
\item We set a fixed size of testing patches to be $1000\times 1000$ pixels that correspond to $0.5$mm $\times 0.5$mm.
\item We ignored background pixels (patches) by setting them to \emph{white}. We performed this by analyzing both homogeneity and gradient change of each patch whereas a threshold was used to exclude background patches (which are widely homogenous and do not exhibit much gradient information).
\item We manually selected $n_i$ patches per WSI with $i=\{1,2,\dots,24\}$. The visual patch selection aimed to extract a small number of patches that represent all dominant tissue textures in each WSI (in fact, every scan does contain multiple texture patterns).
\item Each selected patch was then removed from the scan and saved separately as a testing patch. 
\item The remaining parts of the WSI can be used to construct a training dataset.
\end{enumerate}

\begin{figure*}[htbp]
\begin{center}
\includegraphics[width=0.37\textwidth]{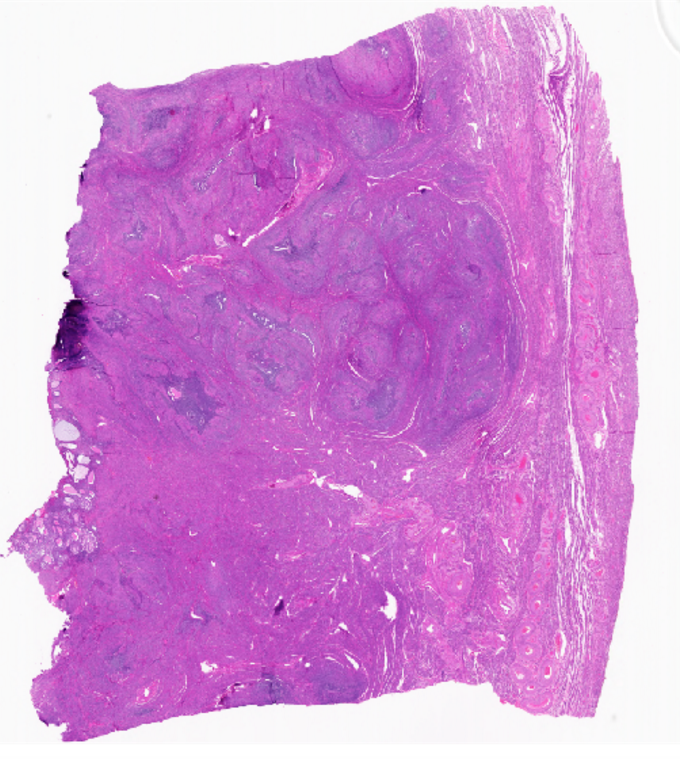}
\includegraphics[width=0.37\textwidth]{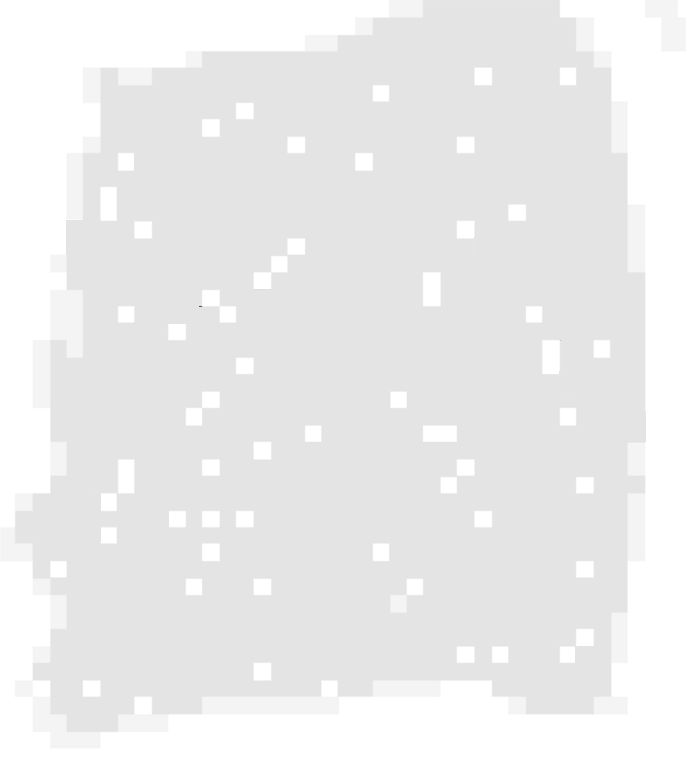}
\includegraphics[width=0.66\textwidth]{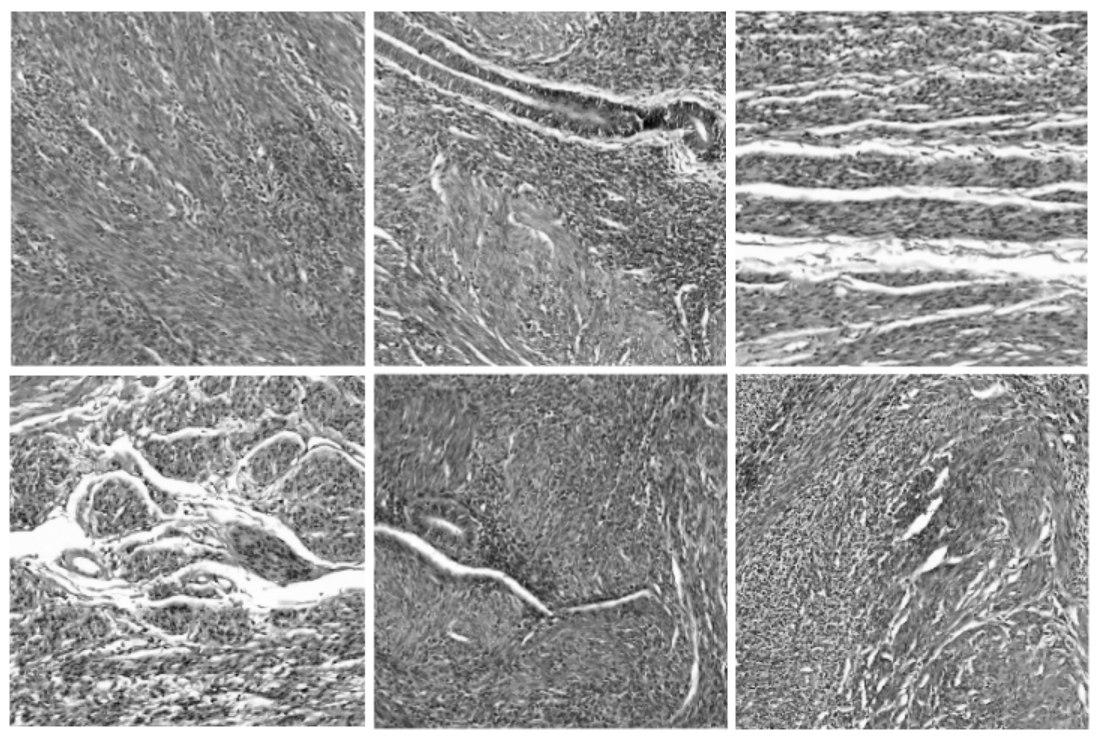}
\caption{A whole scan image (top left) and the visualization of its patch selection (top right). White squares are selected for testing. The grey region can be used to generate training/indexing dataset. Six samples, 1000 $\times$ 1000 and grey-scaled, are displayed as examples for testing patches extracted from the whole scan. }
\label{fig:WSIpatch_selection}
\end{center}
\end{figure*}

Fig. \ref{fig:WSIpatch_selection} demonstrate the patch selection for a sample WSI. The scans are available online and can be downloaded \footnote{Downloading the dataset: \url{http://kimia.uwaterloo.ca}}. The dimensions and number of testing patches for each scan are reported in Table \ref{tab:scannum}.

\begin{figure*}[htbp]
\begin{center}
\fcolorbox{black}{black}{\includegraphics[width=4cm, height=4cm]{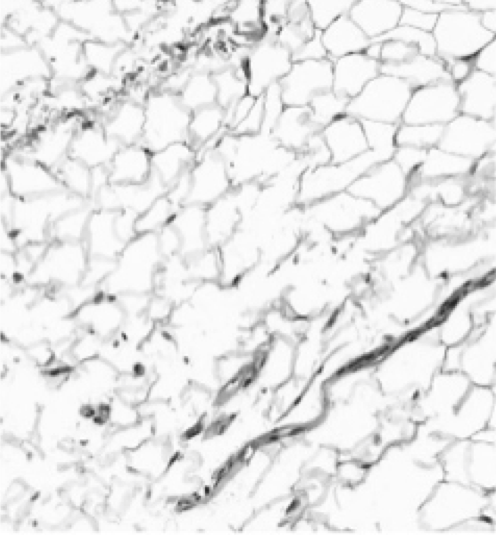}}
\includegraphics[width=4cm, height=4cm]{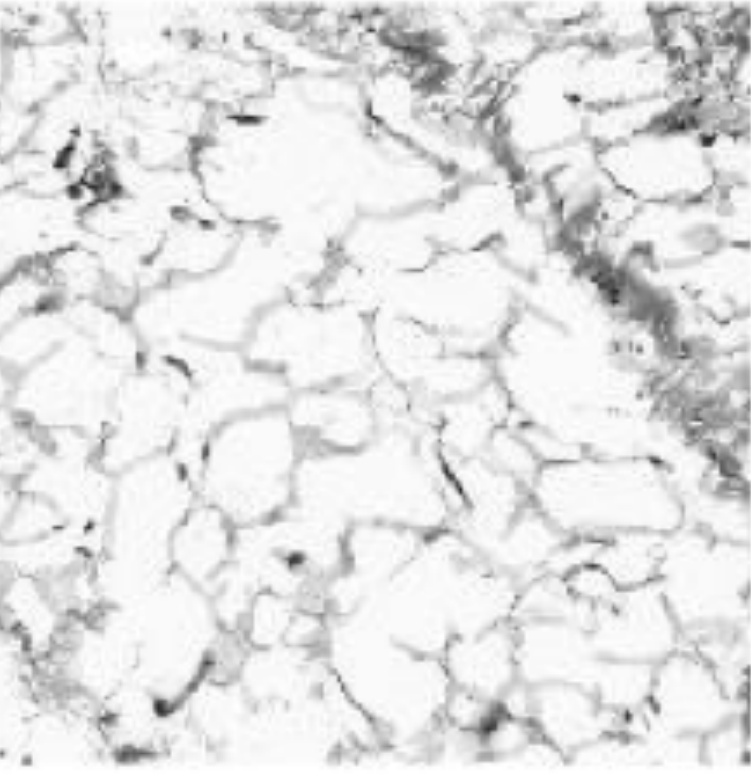}
\includegraphics[width=4cm, height=4cm]{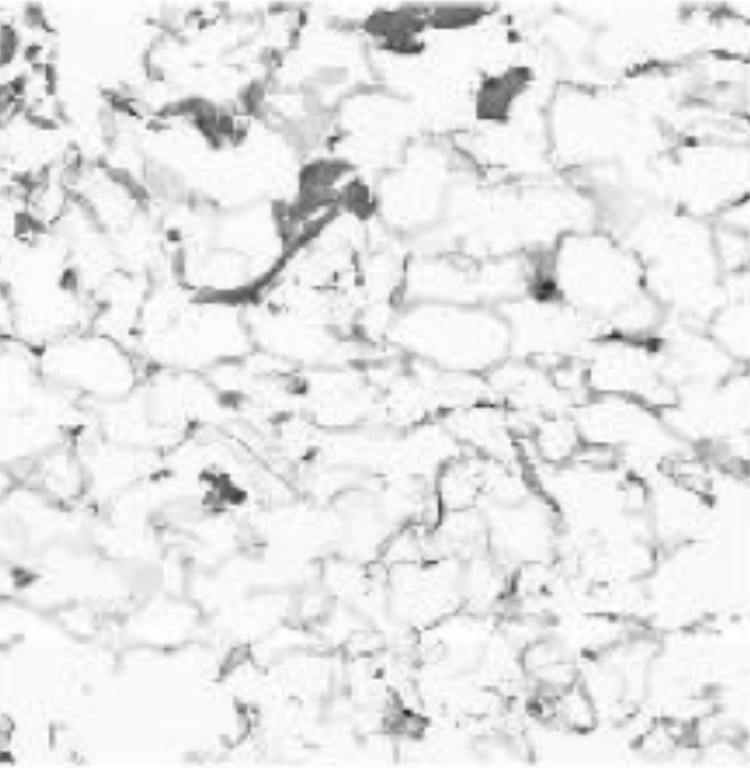}
\includegraphics[width=4cm, height=4cm]{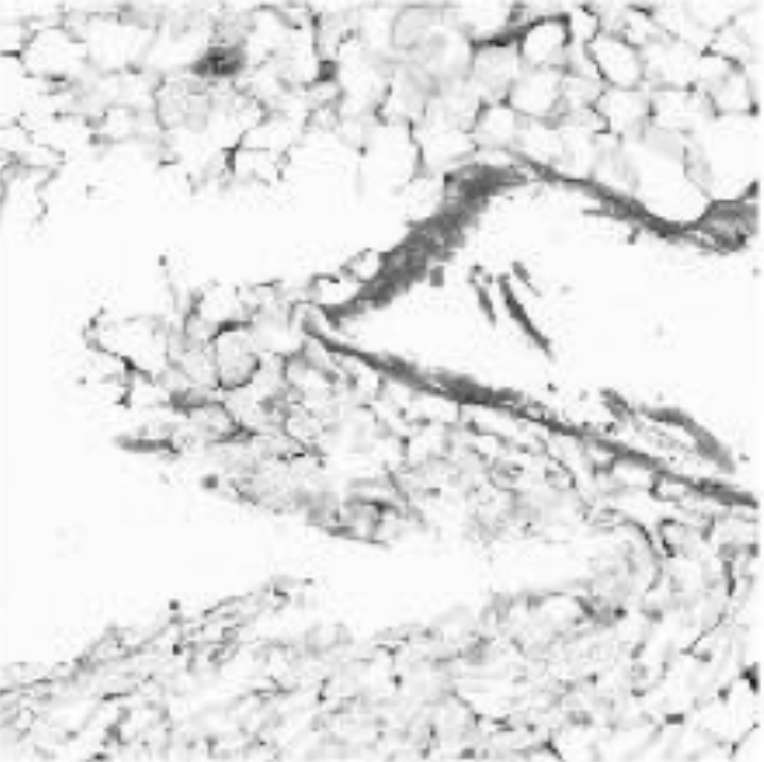}
\caption{Sample for retrieval: a query patch (left image, framed) and its top three search results. }
\label{fig:QueryRetrieval}
\end{center}
\end{figure*}

\begin{table}[]
  \caption{\emph{Kimia Path24} dataset: Properties of each scan}
  \label{tab:scannum}
  \centering
  \begin{tabular}{@{}ccc@{}}
    \toprule
    Scan Index & Dimensions & Number of Test Patches   \\ \toprule
    0 & 40300 $\times$ 58300 & 65\\
    1 & 37800 $\times$ 50400 & 65\\
    2 & 44600 $\times$ 77800 & 65\\
    3 & 50100 $\times$ 77200 & 75\\
    4 & 26500 $\times$ 13600 & 15\\
    5 & 27800 $\times$ 32500 & 40\\
    6 & 38300 $\times$ 51800 & 70\\
    7 & 29600 $\times$ 34300 & 50\\
    8 & 40100 $\times$ 41500 & 60\\
    9 & 40000 $\times$ 50700 & 60\\
    10 & 47500 $\times$ 84700 & 70\\
    11 & 44100 $\times$ 52700 & 70\\
    12 & 45400 $\times$ 60100 & 70\\
    13 & 79900 $\times$ 56600 & 60\\
    14 & 42800 $\times$ 58200 & 60\\
    15 & 20200 $\times$ 57100 & 30\\
    16 & 35300 $\times$ 46300 & 45\\
    17 & 48700 $\times$ 61500 & 45\\
    18 & 26000 $\times$ 49600 & 25\\
    19 & 30700 $\times$ 70400 & 25\\
    20 & 48200 $\times$ 81400 & 65\\
    21 & 38500 $\times$ 40500 & 65\\
    22 & 40500 $\times$ 45700 & 65\\
    23 & 36900 $\times$ 49000 & 65\\ \bottomrule
  \end{tabular}
\end{table}

\subsection{Accuracy Calculation}
We have a total of $n_\textrm{tot}=1,325$ patches $P_{s}^j$ that belong to 24 sets $\Gamma_s=\{P_{s}^i | s\in S, i=1,2\dots,n_{\Gamma_s}\}$ with $s=0,1,2,\dots,23$ ($n_{\Gamma_s}$ reported in the last column of Table \ref{tab:scannum}). Looking at the set of retrieved images $R$ for any experiment, the \textbf{patch-to-scan accuracy} $\eta_\textrm{p}$ can be given as
\begin{equation}
\label{eqn:etap}
\eta_\textrm{p}=\frac{\sum\limits_{s\in S} |R \cap \Gamma_s|}{n_{tot}}.
\end{equation}
As well, we calculate the \textbf{whole-scan accuracy} $\eta_\textrm{W}$ as
\begin{equation}
\label{eqn:etaw}
\eta_\textrm{W}=\frac{1}{24} \sum\limits_{s\in S}\frac{|R \cap \Gamma_s|}{n_{\Gamma_s}}.
\end{equation}
Hence, the total accuracy $\eta_\textrm{total}$ can be defined to take into account both patch-to-scan and whole-scan accuracies: 
\begin{equation}
\label{eqn:etat}
\eta_\textrm{total}= \eta_\textrm{p}\times\eta_\textrm{W}.
\end{equation}
The Matlab and Python code for accuracy calculations can also be downloaded.

%

\section{Experiments on ``Kimia Path24'' Dataset}
To provide preliminary results to set a benchmark line for the proposed data set, we performed three series of retrieval and classification experiments: (\textit{i}) Local Binary Pattern (LBP), (\textit{ii}) Bag of Words (BOW), and (\textit{iii})  Convolutional Neural Networks (CNN). The following subsections will elaborate on every series of experiments and report the overall accuracy of each method. Table \ref{tab:table_experiment} provides the results for all tested methods.

\subsection{Experiment 1: Local Binary Patterns}
For the first experiment, LBP histograms are computed by setting several different parameters to identify the proper windows size. In particular, the LBP is designed using MATLAB 2016b according to \cite{ojala2002multiresolution}. After computation of descriptors for test and training data for each configuration, we evaluate the discriminative power of each one by $k$-NN search (with $k=1$) to find similar patches. Test images with actual sizes ($1000\times1000$) are fed to the LBP operator. For the training set, as discussed in the \ref{sec:dataset}, we have the option to create our training data from each scan depending on retrieval or classification method. The scans are tiled into patches of the same size as the test patches without any overlap. Then we remove patches with a homogeneity of more than 99\%. Using this configuration, we extracted $27,055$ training images which are compared against each test patch for similarity measurement. Non-rotational versions of uniform LBP \cite{ojala2002multiresolution} of radii 1, 2 and 3 with 8,16 and 24 neighbors, respectively, are applied to create a histogram of length 59, 243 and 555. Varying radius helps LBP to capture the texture in different scales, which could be a significant research question in WSI processing \cite{gilbertson2006primary}. As shown in Table \ref{tab:table_experiment},  a longer radius contributes to an increase in both accuracy (2\%-3\% at most) and descriptor length.

\subsection{Experiment 2: Bag-of-Words Approach}
For the second experiment, a dictionary learning approach, i.e., Bag-of-Words (BOW), was designed by minimizing the error of the training samples to build the over-complete dictionary (i.e., \emph{the codebook}) \cite{joachims1998text, mccallum1998comparison}. The design of the BOW approach was conducted according to \cite{lazebnik2006beyond}. The frequency of occurrence of the word histogram can be assigned to describe the input based on the trained dictionary. To train the pathology patch dictionary, two descriptors are used: the raw pixels and the LBP of the raw pixels in each neighborhood. Before extracting the descriptors, 300 patches are selected randomly from all available patches such that their gradient is larger than the average gradient value of all patches for each scan. To accelerate the dictionary training, all 7,200 $=24\times 300$ patches are down-sampled to 500$\times$500 and meshed into 16$\times$16 grids without overlap. Raw pixel descriptors and LBP features are extracted from these sub-patches. The extracted descriptors are used for training the dictionary whose size is set to 800. The word-frequency histogram of each patch is then encoded using the learned dictionary. Finally, the word histogram with its corresponding scan labels is fed to SVM classification with histogram intersection kernel \cite{CC01a}. From the obtained results, depicted in Table \ref{tab:table_experiment}, we can see that using raw pixels or LBP features result in roughly the same accuracy. However, LBP was observed to be much quicker to train and computationally less expensive as compared to raw pixels.

\subsection{Experiment 3: Convolutional Neural Network}
For the last experiment, we used two deep convolution neural networks (CNN), architecturally and conceptually inspired by Alexnet \cite{krizhevsky2012imagenet} and VGG16
\cite{simonyan2014very}. The first network, CNN$_1$, is trained from scratch using end-to-end training with limited pre-processing. For each of the 24 scans, patches are extracted with 40\% overlap, and 99\% homogeneity threshold, further filtered with top 60\% of
gradient values and re-sized to 128 $\times$ 128 using bicubic interpolation. This results in a total of 40,513 patches for training. The CNN consists of 3
convolutional layers with 3 $\times$ 3 kernels, each with 2$\times$ 2 max-pooling with 64, 128, and 256 filters,  respectively. The output from the last convolution layer is fed into a fully-connected layer with 1,024 neurons and subsequently to 24 units with softmax activation for classification. Other than the last layer, all other layers in
the network use \emph{ReLu} as the activation function. The Adam optimizer is used with a learning rate of $0.001$ and the categorical cross-entropy is used as the loss
function. The network achieves the highest accuracy of $\eta_\textrm{total}= 41.80\%$ for the proposed dataset.

In another experiment, a slightly different architecture, CNN$_2$ was considered for classifying the images. A shallow structure of three blocks with 16, 32, and 64 filters was investigated. The high-level extracted features from the last convolutional layer are passed to one fully connected layer with 625 neurons, followed by a softmax classifier at the top.  Our experiments showed that applying 35\% dropout with $L_2$ norm regularization technique achieves around 40.24\% accuracy.  

\begin{table}[t]
\caption{Preliminary results for the Kimia Path24 dataset. LBP uses different distance measures ($L_1,L_2$ and $\chi^2$) in different configurations $(n,r)$ for $n$ neighbours and  radius $r$. Best results for $\eta_p,\eta_W$ and $\eta_\textrm{total}$ are highlighted in bold. }
\label{tab:table_experiment}
\centering
\begin{tabular}{@{}lccc@{}}
\toprule
\textbf{Method}          & $\eta_{p}$    & $\eta_{W}$    & $\eta_{total}$ \\ \toprule
LBP$^{u}_{(8,1)},\chi^2$  & $62.49$ & $58.92$ & $36.82$  \\
LBP$^{u}_{(8,1)},L_1$   & $61.13$ & $57.50$ & $35.15$  \\
LBP$^{u}_{(8,1)},L_2$   & $56.45$ & $52.95$ & $29.89$  \\
LBP$^{u}_{(16,2)},\chi^2$ & $63.62$ & $59.51$ & $37.86$  \\
LBP$^{u}_{(16,2)},L_1$  & $62.26$ & $58.19$ & $36.23$  \\
LBP$^{u}_{(16,2)},L_2$  & $55.77$ & $52.12$ & $29.07$  \\
LBP$^{u}_{(24,3)},\chi^2$ & $64.67$ & $61.08$ & $39.50$  \\
LBP$^{u}_{(24,3)},L_1$  & $\boldsymbol{66.11}$ & $62.52$ & $41.33$  \\
LBP$^{u}_{(24,3)},L_2$  & $59.01$ & $55.94$ & $33.01$  \\ \hline
BOW$_\textrm{RAW}$        & $64.98$    & $61.02$    & $39.65$     \\
BOW$_\textrm{LBP}$        & $64.68$    & $59.33$    & $38.37$     \\ \hline
CNN$_1$     & $64.98$    & $\boldsymbol{64.75}$    & $\boldsymbol{41.80}$  \\
CNN$_2$     & $64.98$    & $61.92$    & $40.24$  \\ \bottomrule
\end{tabular}
\end{table} 

\subsection{Analysis and Discussion}

The data set we are proposing may be regarded \emph{easy} because we are trying to match patches that come from the same scan/patient. However, as the results demonstrate, this is clearly not the case. The BoW approach was able to provide a much more compact representation (codebook). Besides, the storage requirements are quite low. However, the geometric and spatial relations between words are generally neglected. Apparently, the outliers in the training data make it fail to perform properly. CNN achieved the best result of $41.80\%$ followed by LBP with an accuracy of $41.33\%$. CNN, most likely, can still improve if a larger training set is extracted from the scans. 
And as for LBP, perhaps using some classifier may increase the accuracy. One should bear in mind that LBP did in fact process the images in their original dimensions whereas CNNs required substantial downsampling.  Taking into account the training complexity, one may prefer LBP over CNNs for the former is intrinsically simple and fast. As well, LBP has been designed to deal with textures at the spatial level.

\section{Summary}
In this paper, we put forward Kimia Path24, a new data set for retrieval and classification of digital pathology scans. We selected 24 scans from a large pool of scans through visual inspection. The criterion was to select \emph{texturally different} images. Hence, the proposed data set is rather a computer vision data set (as in contrast to a pathological data set) because visual attention has been spent on the diversity of patterns and not on anatomies and malignancies. The task, hence, is whether machine-learning algorithms can discriminate among diverse patterns. To start a baseline, we applied LBP, the dictionary approach and CNNs to classify patches. The results show that the task is apparently quite difficult.

\vspace{0.1in}
The data set can be downloaded from authors' website: \url{http://kimia.uwaterloo.ca} 

{\small
\bibliographystyle{ieee}
\bibliography{egbib}

\begin{thebibliography}{10}\itemsep=-1pt

\bibitem{akakin2012content}
H.~C. Akakin and M.~N. Gurcan.
\newblock Content-based microscopic image retrieval system for multi-image
  queries.
\newblock {\em IEEE transactions on information technology in biomedicine},
  16(4):758--769, 2012.

\bibitem{Janabi2011}
S.~Al-Janabi, A.~Huisman, and P.~J. Van~Diest.
\newblock Digital pathology: current status and future perspectives.
\newblock {\em Histopathology}, 61(1):1--9, 2012.

\bibitem{icpram17}
M.~Babaie, H.~R. Tizhoosh, S.~Zhu, and M.~E. Shiri.
\newblock Retrieving similar x-ray images from big image data using radon
  barcodes with single projections.
\newblock In {\em Proceedings of the 6th International Conference on Pattern
  Recognition Applications and Methods}, pages 557--566, 2017.

\bibitem{bankhead2017qupath}
P.~Bankhead, M.~B. Loughrey, J.~A. Fern{\'a}ndez, Y.~Dombrowski, D.~G. McArt,
  P.~D. Dunne, S.~McQuaid, R.~T. Gray, L.~J. Murray, H.~G. Coleman, et~al.
\newblock Qupath: Open source software for digital pathology image analysis.
\newblock {\em bioRxiv}, page 099796, 2017.

\bibitem{barker2016automated}
J.~Barker, A.~Hoogi, A.~Depeursinge, and D.~L. Rubin.
\newblock Automated classification of brain tumor type in whole-slide digital
  pathology images using local representative tiles.
\newblock {\em Medical image analysis}, 30:60--71, 2016.

\bibitem{bartels1992bayesian}
P.~Bartels, D.~Thompson, M.~Bibbo, and J.~Weber.
\newblock Bayesian belief networks in quantitative histopathology.
\newblock {\em Analytical and quantitative cytology and histology/the
  International Academy of Cytology [and] American Society of Cytology},
  14(6):459--473, 1992.

\bibitem{caicedo2011content}
J.~C. Caicedo, F.~A. Gonz{\'a}lez, and E.~Romero.
\newblock Content-based histopathology image retrieval using a kernel-based
  semantic annotation framework.
\newblock {\em Journal of biomedical informatics}, 44(4):519--528, 2011.

\bibitem{carpenter2012call}
A.~E. Carpenter, L.~Kamentsky, and K.~W. Eliceiri.
\newblock A call for bioimaging software usability.
\newblock {\em Nature methods}, 9(7):666, 2012.

\bibitem{CC01a}
C.-C. Chang and C.-J. Lin.
\newblock {LIBSVM}: A library for support vector machines.
\newblock {\em ACM Transactions on Intelligent Systems and Technology},
  2:1--27, 2011.

\bibitem{della2009eslide}
V.~Della~Mea, N.~Bortolotti, and C.~Beltrami.
\newblock eslide suite: an open source software system for whole slide imaging.
\newblock {\em Journal of clinical pathology}, 62(8):749--751, 2009.

\bibitem{Gabril2010}
M.~Y. Gabril and G.~Yousef.
\newblock Informatics for practicing anatomical pathologists: marking a new era
  in pathology practice.
\newblock {\em Mod Pathol}, 23(3):349--358, 2010.

\bibitem{ghaznavi2013digital}
F.~Ghaznavi, A.~Evans, A.~Madabhushi, and M.~Feldman.
\newblock Digital imaging in pathology: whole-slide imaging and beyond.
\newblock {\em Annual Review of Pathology: Mechanisms of Disease}, 8:331--359,
  2013.

\bibitem{gilbertson2006primary}
J.~R. Gilbertson, J.~Ho, L.~Anthony, D.~M. Jukic, Y.~Yagi, and A.~V. Parwani.
\newblock Primary histologic diagnosis using automated whole slide imaging: a
  validation study.
\newblock {\em BMC clinical pathology}, 6(1):4, 2006.

\bibitem{gurcan2009review}
M.~N. Gurcan, L.~E. Boucheron, A.~Can, A.~Madabhushi, N.~M. Rajpoot, and
  B.~Yener.
\newblock Histopathological image analysis: A review.
\newblock {\em IEEE reviews in biomedical engineering}, 2:147--171, 2009.

\bibitem{gutman2013cancer}
D.~A. Gutman, J.~Cobb, D.~Somanna, Y.~Park, F.~Wang, T.~Kurc, J.~H. Saltz,
  D.~J. Brat, L.~A. Cooper, and J.~Kong.
\newblock Cancer digital slide archive: an informatics resource to support
  integrated in silico analysis of tcga pathology data.
\newblock {\em Journal of the American Medical Informatics Association},
  20(6):1091--1098, 2013.

\bibitem{hamilton1994expert}
P.~Hamilton, N.~Anderson, P.~Bartels, and D.~Thompson.
\newblock Expert system support using bayesian belief networks in the diagnosis
  of fine needle aspiration biopsy specimens of the breast.
\newblock {\em J. of Clinical Pathology}, 47(4):329--336, 1994.

\bibitem{ho2006use}
J.~Ho, A.~V. Parwani, D.~M. Jukic, Y.~Yagi, L.~Anthony, and J.~R. Gilbertson.
\newblock Use of whole slide imaging in surgical pathology quality assurance:
  design and pilot validation studies.
\newblock {\em Human pathology}, 37(3):322--331, 2006.

\bibitem{joachims1998text}
T.~Joachims.
\newblock Text categorization with support vector machines: Learning with many
  relevant features.
\newblock {\em Machine learning: ECML-98}, pages 137--142, 1998.

\bibitem{krizhevsky2012imagenet}
A.~Krizhevsky, I.~Sutskever, and G.~E. Hinton.
\newblock Imagenet classification with deep convolutional neural networks.
\newblock In {\em Advances in neural information processing systems}, pages
  1097--1105, 2012.

\bibitem{lazebnik2006beyond}
S.~Lazebnik, C.~Schmid, and J.~Ponce.
\newblock Beyond bags of features: Spatial pyramid matching for recognizing
  natural scene categories.
\newblock In {\em Computer vision and pattern recognition, 2006 IEEE computer
  society conference on}, volume~2, pages 2169--2178. IEEE, 2006.

\bibitem{lecun1998gradient}
Y.~LeCun, L.~Bottou, Y.~Bengio, and P.~Haffner.
\newblock Gradient-based learning applied to document recognition.
\newblock {\em Proceedings of the IEEE}, 86(11):2278--2324, 1998.

\bibitem{lowe1999object}
D.~G. Lowe.
\newblock Object recognition from local scale-invariant features.
\newblock In {\em Computer vision, 1999. The proceedings of the seventh IEEE
  international conference on}, volume~2, pages 1150--1157. Ieee, 1999.

\bibitem{Madabhushi2017}
A.~Madabhushi and G.~Lee.
\newblock Image analysis and machine learning in digital pathology: Challenges
  and opportunities.
\newblock {\em Medical Image Analysis}, 33:170--175, 2017.

\bibitem{masood2009texture}
K.~Masood and N.~Rajpoot.
\newblock Texture based classification of hyperspectral colon biopsy samples
  using clbp.
\newblock In {\em IEEE International Symposium on Biomedical Imaging}, pages
  1011--1014, 2009.

\bibitem{mccallum1998comparison}
A.~McCallum, K.~Nigam, et~al.
\newblock A comparison of event models for naive bayes text classification.
\newblock In {\em AAAI-98 workshop on learning for text categorization}, volume
  752, pages 41--48. Citeseer, 1998.

\bibitem{mehta2009content}
N.~Mehta, A.~Raja'S, and V.~Chaudhary.
\newblock Content based sub-image retrieval system for high resolution
  pathology images using salient interest points.
\newblock In {\em IEEE International Conference of the Engineering in Medicine
  and Biology Society}, pages 3719--3722, 2009.

\bibitem{nanni2010local}
L.~Nanni, A.~Lumini, and S.~Brahnam.
\newblock Local binary patterns variants as texture descriptors for medical
  image analysis.
\newblock {\em Artificial intelligence in medicine}, 49(2):117--125, 2010.

\bibitem{ojala1994performance}
T.~Ojala, M.~Pietikainen, and D.~Harwood.
\newblock Performance evaluation of texture measures with classification based
  on kullback discrimination of distributions.
\newblock In {\em Proceedings of the 12th IAPR International Conference on
  Computer Vision \& Image Processing}, volume~1, pages 582--585, 1994.

\bibitem{ojala2002multiresolution}
T.~Ojala, M.~Pietikainen, and T.~Maenpaa.
\newblock Multiresolution gray-scale and rotation invariant texture
  classification with local binary patterns.
\newblock {\em IEEE Transactions on pattern analysis and machine intelligence},
  24(7):971--987, 2002.

\bibitem{pantanowitz2011review}
L.~Pantanowitz, P.~N. Valenstein, A.~J. Evans, K.~J. Kaplan, J.~D. Pfeifer,
  D.~C. Wilbur, L.~C. Collins, T.~J. Colgan, et~al.
\newblock Review of the current state of whole slide imaging in pathology.
\newblock {\em Journal of pathology informatics}, 2(1):36, 2011.

\bibitem{reyad2014comparison}
Y.~A. Reyad, M.~A. Berbar, and M.~Hussain.
\newblock Comparison of statistical, lbp, and multi-resolution analysis
  features for breast mass classification.
\newblock {\em Journal of medical systems}, 38(9):100, 2014.

\bibitem{sertel2009computer}
O.~Sertel, J.~Kong, H.~Shimada, U.~Catalyurek, J.~H. Saltz, and M.~N. Gurcan.
\newblock Computer-aided prognosis of neuroblastoma on whole-slide images:
  Classification of stromal development.
\newblock {\em Pattern recognition}, 42(6):1093--1103, 2009.

\bibitem{simonyan2014very}
K.~Simonyan and A.~Zisserman.
\newblock Very deep convolutional networks for large-scale image recognition.
\newblock {\em arXiv preprint arXiv:1409.1556}, 2014.

\bibitem{tashk2013automatic}
A.~Tashk, M.~S. Helfroush, H.~Danyali, and M.~Akbarzadeh.
\newblock An automatic mitosis detection method for breast cancer
  histopathology slide images based on objective and pixel-wise textural
  features classification.
\newblock In {\em Conference on Information and Knowledge Technology}, pages
  406--410, 2013.

\bibitem{weind1998invasive}
K.~L. Weind, C.~F. Maier, B.~K. Rutt, and M.~Moussa.
\newblock Invasive carcinomas and fibroadenomas of the breast: comparison of
  microvessel distributions--implications for imaging modalities.
\newblock {\em Radiology}, 208(2):477--483, 1998.

\bibitem{weinstein2009overview}
R.~S. Weinstein, A.~R. Graham, L.~C. Richter, G.~P. Barker, E.~A. Krupinski,
  A.~M. Lopez, K.~A. Erps, A.~K. Bhattacharyya, Y.~Yagi, and J.~R. Gilbertson.
\newblock Overview of telepathology, virtual microscopy, and whole slide
  imaging: prospects for the future.
\newblock {\em Human pathology}, 40(8):1057--1069, 2009.

\bibitem{williams2010telepathology}
S.~Williams, W.~H. Henricks, M.~J. Becich, M.~Toscano, and A.~B. Carter.
\newblock Telepathology for patient care: what am i getting myself into?
\newblock {\em Advances in anatomic pathology}, 17(2):130--149, 2010.

\bibitem{zhang2015towards}
X.~Zhang, W.~Liu, M.~Dundar, S.~Badve, and S.~Zhang.
\newblock Towards large-scale histopathological image analysis: Hashing-based
  image retrieval.
\newblock {\em IEEE Transactions on Medical Imaging}, 34(2):496--506, 2015.

\bibitem{zheng2003design}
L.~Zheng, A.~W. Wetzel, J.~Gilbertson, and M.~J. Becich.
\newblock Design and analysis of a content-based pathology image retrieval
  system.
\newblock {\em IEEE Transactions on Information Technology in Biomedicine},
  7(4):249--255, 2003.

\end{thebibliography}
}

\end{document}